\documentclass{article}
\usepackage{spconf,amsmath,graphicx,hyperref}

\usepackage{xcolor}
\usepackage{multirow}
\definecolor{new_blue}{RGB}{0,176,240}
\definecolor{new_purple}{RGB}{112,48,160}
\usepackage{booktabs}

\title{Pseudo-Siamese Network for Planning in Target-Oriented Proactive Dialogues}
\name{Xinyue Kang\textsuperscript{1,2}*, Maodong Li\textsuperscript{1,2}*\thanks{*Equal contribution}, Yibin Zheng\textsuperscript{1,2}, Fang Kong\textsuperscript{1,2}†\thanks{†Corresponding author}\thanks{This work was supported by the Project 62276178 under the National Natural Science Foundation of China, the Key Project 23KJA520012 under the Natural Science Foundation of Jiangsu Higher Education Institutions, the project 22YJCZH091 of Humanities and Social Science Fund of Ministry of Education and the Priority Academic Program Development of Jiangsu Higher Education Institutions. They are also with Jiangsu Key Lab of Language Computing, Suzhou, 215021, P. R. China.}}
\address{\textsuperscript{1}School of Computer Science and Technology, Soochow University \\
\textsuperscript{2}Jiangsu Key Lab of Language Computing, Suzhou 215123, China \\
\{20244227013@stu, 20254027002@stu, 20234227027@stu, kongfang@\}suda.edu.cn
}

\begin{document}
\ninept
\maketitle
\begin{abstract}
A target-oriented proactive dialogue system is designed to steer conversations toward predefined targets while actively providing suggestions. The core paradigm of such a system is to plan a reasonable dialogue path and subsequently guide language models (e.g., pre-trained or large language models) to generate responses, where dialogue path planning serves as the central component—a novel yet under-explored problem. In this work, we propose a \textbf{F}orward-\textbf{F}ocused \textbf{B}idirectional \textbf{P}seudo-\textbf{S}iamese \textbf{N}etwork (FF-BPSN) for dialogue path planning toward predefined dialogue targets. FF-BPSN employs two identical transformer-based decoders for forward and backward planning, together with a forward-focused module that integrates bidirectional information to construct the final forward path. This path benefits from bidirectional planning while prioritizing forward information. We then employ the planned path to guide language models in response generation. Extensive experiments on DuRecDial and DuRecDial 2.0 demonstrate that FF-BPSN achieves state-of-the-art performance in dialogue path planning and significantly enhances the effectiveness of target-oriented proactive dialogue systems.
\end{abstract}
\section{Introduction}
\label{sec:intro}
\begin{figure}[t]
    \centering
    \includegraphics[width=0.48\textwidth]{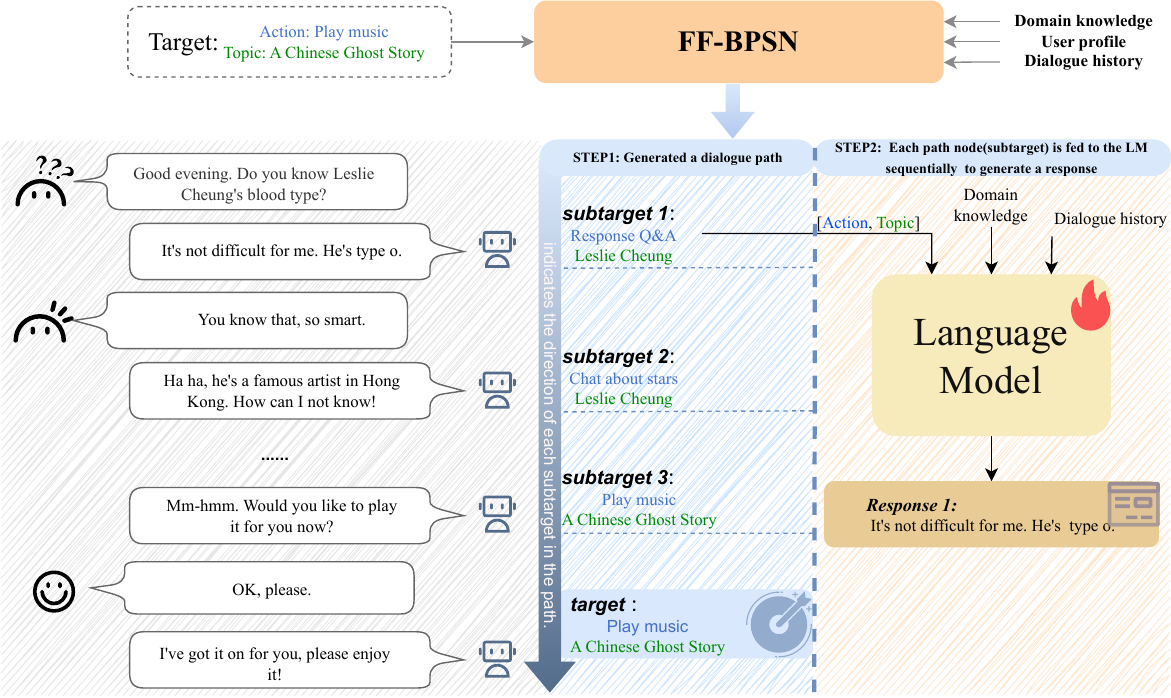}
    \caption{
    An example of a target-oriented proactive dialogue system. Blue denotes Action, and green denotes Topic. The path is generated by FF-BPSN based on the dialogue target, domain knowledge, user profile, and dialogue history. Each dialogue turn is then generated conditioned on the current subtarget, domain knowledge, and dialogue history.
    }
    \label{fig1}
\end{figure}

A target-oriented proactive dialogue system is designed to guide conversations toward predefined targets while proactively offering suggestions when appropriate \cite{lin2024screen,wang2023target}. Ensuring coherence across multiple turns is crucial for effectiveness, and such systems have attracted increasing research interest in recent years \cite{ijcai2023p738,deng2025proactive,dong-etal-2025-protod}. Building on prior work \cite{liu-etal-2020-towards-conversational,wang-etal-2023-dialogue,wang2024target}, the dominant paradigm involves first planning a dialogue path and then guiding pre-trained language models (PLMs) or large language models (LLMs) to generate responses. Dialogue path planning lies at the core of this paradigm yet remains novel and under-explored. In this paper, we follow this paradigm by decomposing the task into two steps (Figure~\ref{fig1}): (i) planning a path composed of [Action, Topic] pairs, and (ii) guiding PLMs or LLMs with the path to produce responses. Each pair serves as a subtarget, with the final pair corresponding to the predefined target. For example, in Figure~\ref{fig1}, the path [Respond Q\&A, Leslie Cheung], …, [Play music, A Chinese Ghost Story] indicates that the final target is to play the music A Chinese Ghost Story, with intermediate subtargets such as [Respond Q\&A, Leslie Cheung]. The language models generate responses conditioned on the current subtarget, domain knowledge, and dialogue history, thereby reaching the final target through sequential subtargets.

However, planning a reasonable dialogue path that ensures that dialogues flow naturally remains challenging: (i) the path must be highly relevant to user interests and the dialogue target; (ii) prior work primarily focuses on one-way path planning and does not adequately consider bidirectional planning; and (iii) conversations tend to progress forward, so the forward path should be emphasized. To address these challenges, we propose a \textbf{F}orward-\textbf{F}ocused \textbf{B}idirectional \textbf{P}seudo-\textbf{S}iamese \textbf{N}etwork (FF-BPSN) to plan a dialogue path toward a predefined target. This path is then leveraged to guide PLMs or LLMs in response generation. Specifically, the forward and backward paths are decoded by two transformer-based decoders, forming pseudo-siamese networks. A forward-focused module subsequently integrates information from both paths to generate the final forward path. The resulting forward dialogue path guides PLMs (DialoGPT, BART, GPT-2) and LLMs (LLaMA series) in generating responses. Our contributions are as follows:

(1) To the best of our knowledge, this work is the first to investigate a forward-focused bidirectional network for dialogue path planning in target-oriented proactive dialogue systems.
(2) We propose FF-BPSN for dialogue path planning, which consists of two identical transformer-based decoders and a forward-focused module. The resulting dialogue paths significantly enhance response generation performance.
(3) Extensive experiments show that FF-BPSN achieves state-of-the-art performance in dialogue path planning, leading to substantial improvements in dialogue system performance\footnote{https://github.com/imaodong/FF-BSPN}.

\section{Related Work}
Target-oriented dialogue systems aim to guide conversations toward predefined targets. \cite{tang-etal-2019-target} introduced coarse-grained keywords to steer system responses. \cite{liu2020gochat} and \cite{xu2020knowledge} employed reinforcement learning (RL) to enhance performance by decomposing the task into subtasks. For knowledge graph (KG)-based applications, \cite{zhong2021keyword} integrated an external KG to improve keyword transitions and response retrieval, while \cite{yang2022topkg} proposed a global RL algorithm that plans dialogue paths with a commonsense KG. Target-oriented proactive dialogue systems not only guide conversations toward predefined targets but also provide timely suggestions \cite{lin2024screen,wang2023target}. \cite{liu-etal-2020-towards-conversational} constructed the DuRecDial dataset and proposed a multi-target-driven model for dialogue generation. To unify multi-subgoal dialogue, \cite{zhang2021kers} introduced a knowledge-enhanced, multi-subgoal-driven proactive model. R-Walker \cite{ma-etal-2021-cr} performed tree-structured reasoning over KGs to generate dialogue content. For dialogue path planning, the Brownian bridge stochastic process was applied to model temporal dynamics \cite{wang-etal-2023-dialogue}, while \cite{wang2023target,wang2024target} developed a planning network for this task. However, these methods do not fully leverage bidirectional path information. To address this gap, we propose bidirectional dialogue path planning with a forward-focused module that integrates information from both directions into a final forward path to guide language models in response generation.
\section{Methodology}
\subsection{Task Definition and Notation Explanation}

Let $D=\{(K^i, P^i, C^i), ({ACT}^i_{j=1:T}, {TOP}^i_{j=1:T}), R^i\}_N^{i=1}$ denote the dialogue dataset, where $N$ is the number of conversations and $T$ is the number of [Action, Topic] pairs. Here, $K^i$ represents domain knowledge as subject–relation–object triples, $P^i$ denotes the user profile, $C^i$ is the dialogue history, and $R^i$ is the response. $ACT$ and $TOP$ indicate the “Action” and “Topic” in each pair (Figure~\ref{fig1}). The dialogue path $({ACT}^i_{1:T}, {TOP}^i_{1:T})$ leads to the target $({ACT}^i_T, {TOP}^i_T)$. FF-BPSN generates this path based on $K^i$, $P^i$, $C^i$, and the target, and the resulting forward path is incorporated into the language model at each turn to generate a response.

\subsection{Path Planning Framework}
\begin{figure}[t]
    \centering
    \includegraphics[width=0.48\textwidth]{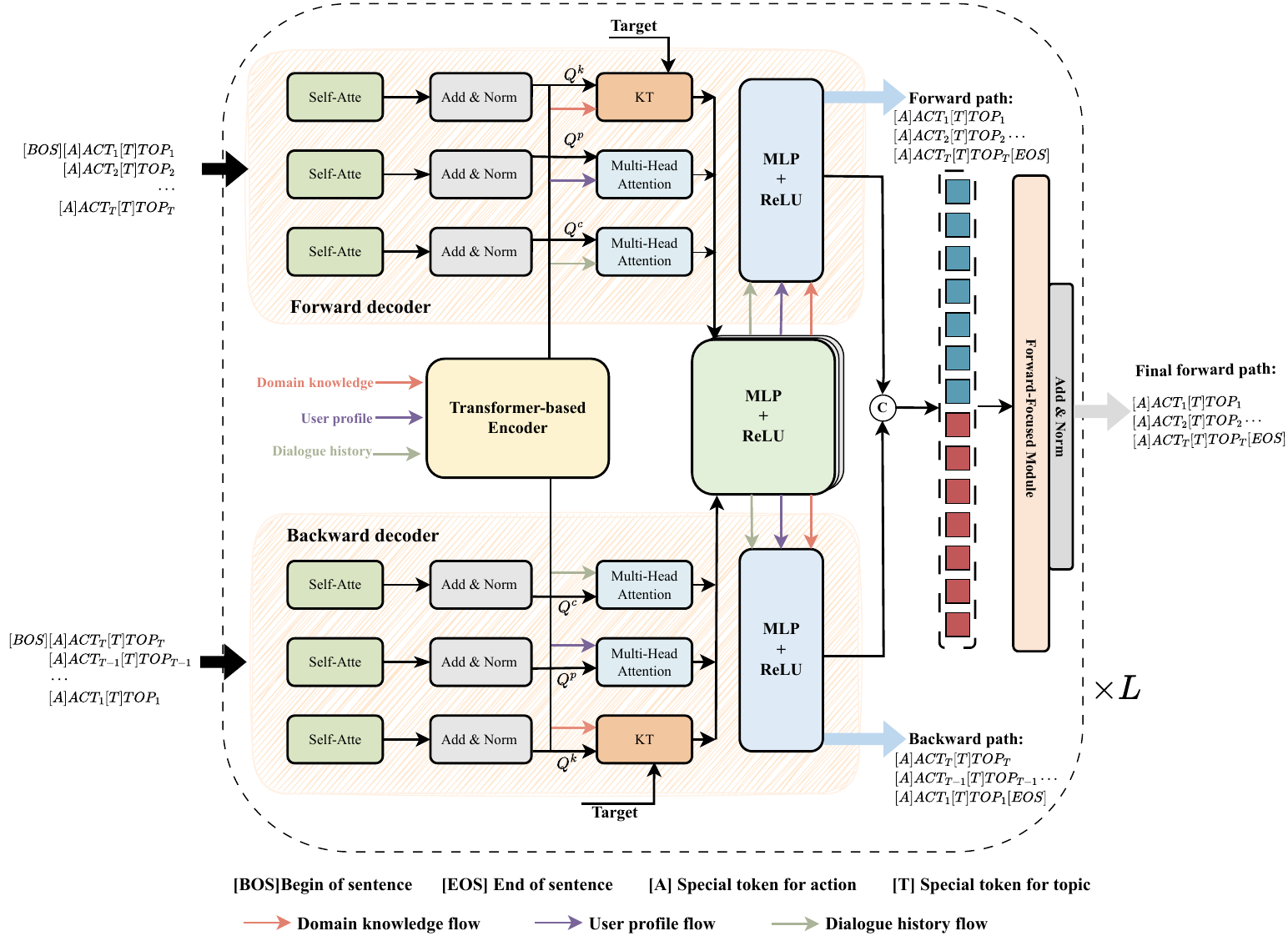}
    \caption{
    Overall architecture of FF-BPSN.
    }
    \label{fig2}
\end{figure}

The path planning framework, built on a Transformer architecture (Figure~\ref{fig2}), generates a dialogue path $({ACT}_{j=1:T}, {TOP}_{j=1:T})$ to ensure smooth transitions and achieve the predefined target. It conducts bidirectional planning—from the current state to the target and vice versa—and merges the results to form the final forward path. We formulate dialogue paths as sequences and introduce two prefix tokens, $[A]$ and $[T]$, to indicate ${ACT}$ and ${TOP}$ during generation (Figure~\ref{fig2}). Domain knowledge is encoded into a graph structure using a Transformer-based encoder, following \cite{wang2023target,wang2024target}, where knowledge is represented as relation–entity pairs and extended by increasing the number of hops from the target node to the current node. Unlike prior work, we directly encode user profile and dialogue history with Transformer-based encoders. The framework input includes domain knowledge $K$, user profile $P$, dialogue history $C$, and the target $({ACT}_T, {TOP}_T)$. FF-BPSN leverages knowledge–target mutual attention (KT), shown effective in \cite{wang2023target,li-etal-2025-multi-hop}. The forward process outputs $[A] {ACT}_1 [T] {TOP}_1 \dots [A] {ACT}_T [T] {TOP}_T$, while the backward process produces the reversed sequence $[A] {ACT}_T [T] {TOP}_T \dots [A] \break {ACT}_1 [T] {TOP}_1$. After bidirectional planning, the forward-focused module generates the final forward path $[A] {ACT}_1 [T] {TOP}_1 \dots \break [A] {ACT}_T [T] {TOP}_T$.


\subsection{Forward-Focused Module}



As shown in Figure~\ref{fig2}, we adopt the standard $ReLU + MLP$ structure for general feature fusion. For the forward-focused component, a forward-focused feature fusion module integrates information from the forward and backward paths to produce the final forward path. Let $F_f$ and $F_b$ denote the forward and backward paths, respectively. The module is formalized as:

\begin{equation}
    O_f = Sigmoid(F^{''}) \cdot F^{''} + (1-Sigmoid(F^{''})) \cdot F^{'}
\end{equation}
\begin{equation}
    F^{''} = MLP([{F}_f;{F}_b])
\end{equation}
\begin{equation}
    F^{'} = {F}_f \cdot F_{weight} + {F}_b \cdot (1-F_{weight})
\end{equation}
\begin{equation}
    F_{weight} = Sigmoid(Linear(F_f))
\end{equation}
where $MLP(\cdot)$ is a vanilla multi-layer perceptron, and $O_f$ denotes the final forward dialogue path output (i.e., the FF-BPSN output). 


\subsection{Training Loss of FF-BPSN}
For the forward, backward, and final forward paths, the losses are denoted by $\mathcal{L}_1$, $\mathcal{L}_2$, and $\mathcal{L}_3$, respectively, and are calculated using cross-entropy. Since the forward and backward paths convey identical meanings, we expect high similarity between them, denoted by $\mathcal{L}_4$. Additionally, we use $\gamma$ and $\beta$ to regulate the influence of forward and backward planning. The total loss function is as follows:

\begin{equation}
\mathcal{L}=\gamma\mathcal{L}_1+\beta\mathcal{L}_2+\mathcal{L}_3+\mathcal{L}_4
\end{equation}
\begin{equation}
    \mathcal{L}_4 = \|F_f-F_b\|_2
\end{equation}


\subsection{Path Decoding and Response Generation}
Through greedy decoding of FF-BPSN, we obtain the final forward path $[A] {ACT}_1 [T] {TOP}_1 \dots [A] {ACT}_T [T] {TOP}_T$. The path planning framework and PLMs (or LLMs) are trained sequentially in a pipeline, with the generated path guiding response generation. At each turn, the current subtarget is extracted from the FF-BPSN path and combined with the dialogue history and domain knowledge to form a natural language prompt, which is then used to fine-tune PLMs and LLMs for response generation.

\begin{equation}
    \hat{R} = LM([C;K;({ACT}_x;{TOP}_x)])
\end{equation}
where $C$ and $K$ denote the dialogue history and domain knowledge, respectively, while ${ACT}_x$ and ${TOP}_x$ represent the current action and topic (i.e., the subtarget). $LM(\cdot)$ denotes a language model, and $\hat{R}$ is the response generated by the language model.

\section{Experiments}


\begin{table}[!t]
\renewcommand{\arraystretch}{0.7}
\small
\centering
\setlength{\tabcolsep}{0.5mm}{
\begin{tabular}{llcccc}
\toprule
\multicolumn{2}{l}{Dataset}            & \#Conversation & \#Response & \#Max. & \#Avg. \\ \midrule
\multirow{3}{*}{DuRecDial}     & Train & 5400    & 84869    & 14     & 7.9    \\
                               & Dev   & 800     & 12508    & 12     & 7.8    \\
                               & Test  & 1804    & 28809    & 13     & 8.0     \\ \midrule
\multirow{3}{*}{DuRecDial 2.0} & Train & 4104    & 66675    & 13     & 8.1    \\
                               & Dev   & 608     & 9677     & 14     & 8.0    \\
                               & Test  & 1368    & 22367    & 13     & 8.2    \\ \bottomrule
\end{tabular}}
\caption{Statistics of datasets.}
\label{Table1}
\end{table}

\begin{table*}[!t]
\small
\renewcommand{\arraystretch}{0.7}
\centering
\setlength{\tabcolsep}{1.0mm}
\begin{tabular}{lccccc|ccccc}
\toprule
              & \multicolumn{5}{c|}{DuRecDial}                               & \multicolumn{5}{c}{DuRecDial 2.0}                            \\ \cmidrule(l){2-11} 
Model          & F1    & BLEU-1/2    & DIST-1/2    & Know. F1 & Succ.  & F1    & BLEU-1/2    & DIST-1/2    & Know. F1 & Succ. \\ \midrule
MGCG\_G       & 33.48 & 0.279/0.203 & 0.007/0.043 & 35.12    & 48.6 & 32.26 & 0.293/0.182 & 0.016/0.051 & 29.35    & 32.2  \\
KERS          & 34.04 & 0.302/0.220 & 0.005/0.030 & 40.75    & 50.5  & 30.11 & 0.282/0.178 & 0.017/0.060 & 33.08    & 40.6  \\
TPC-BART      & 37.22 & 0.338/0.255 & \textbf{0.008}/\textbf{0.083} & 44.52    & 71.5  & 36.28 & 0.296/0.204 & \underline{0.030}/0.093 & 40.22    & 63.6  \\
TPC-GPT       & 41.53 & 0.379/0.301 & 0.007/0.075 & 48.81    & 74.7  & 34.62 & 0.308/0.217 & 0.025/0.082 & 38.80     & 60.7  \\




\midrule
DialoGPT       & 35.85 & 0.344/0.266 & 0.006/0.044 & 37.95    & 49.2  & 39.34 & 0.359/0.266 & 0.015/0.054 & 49.03    & 40.7  \\
DialoGPT w/ FF & 38.65 & 0.357/0.278 & 0.006/0.055 & 45.09    & 61.5  & 44.32 & 0.382/0.287 & 0.017/0.062 & 62.74    & 65.5  \\
BART          & 35.22 & 0.324/0.252 & 0.007/0.066 & 39.77    & 49.9   & 37.20  & 0.285/0.186 & 0.020/0.071 & 45.27    & 52.0    \\
BART w/ FF     & 37.52 & 0.313/0.241 & \textbf{\underline{0.008}}/\underline{0.077} & 44.49    & 58.5  & 40.53 & 0.306/0.207 & 0.020/0.071 & 55.81    & 67.9  \\
GPT-2         & 39.78 & 0.373/0.296 & 0.007/0.060 & 45.28    & 62.0     & 41.62 & 0.375/0.279 & 0.016/0.060 & 54.22    & 50.9  \\ 
GPT-2 w/ FF   & 42.48 & 0.385/0.308 & 0.007/0.071 & 50.47    & 72.1   & 44.74 & 0.391/0.292 & 0.017/0.064 & 64.43    & 64.8  \\ 

LLaMA-1B & 42.55 & 0.397/0.323 & 0.005/0.047 & 48.60 & 61.7 & 40.86 & 0.379/0.271 & 0.021/0.106 & 51.47 & 57.9
\\
LLaMA-1B w/ FF & 46.55 & 0.422/0.349 & 0.006/0.050 & 56.35 & 85.2 & 45.17 & 0.404/0.297 & \textbf{0.023}/\textbf{\underline{0.114}} & 66.91 & 73.7
\\
LLaMA-3B & 43.19 & 0.404/0.328 & 0.006/0.055 & 50.44 & 65.1 & 42.34 & 0.394/0.288 & 0.016/0.065 & 52.80 & 61.4
\\
LLaMA-3B w/ FF & \textbf{\underline{47.43}} & \textbf{\underline{0.429}}/\textbf{\underline{0.353}} & 0.006/0.057 & \textbf{\underline{57.34}} & \textbf{\underline{86.4}} & \textbf{\underline{45.84}} & \textbf{\underline{0.408}}/\textbf{\underline{0.299}} & 0.018/0.083 &\textbf{\underline{67.27}} & \textbf{\underline{79.8}}
\\
\bottomrule
\end{tabular}
\caption{Main results. \textbf{Bold} indicates the best results among category models w/ FF, while \underline{underlined} marks the best results across all models.}
\label{Table2}
\end{table*}

\begin{table}[!t]
\renewcommand{\arraystretch}{0.7}
\small
\centering
\setlength{\tabcolsep}{0.5mm}{
\begin{tabular}{lcccc|cccc}
\toprule
                       & \multicolumn{4}{c|}{DuRecDial}                                            & \multicolumn{4}{c}{DuRecDial 2.0}                                        \\ \cmidrule(l){2-9} 
\multirow{2}{*}{Model} & \multicolumn{2}{c}{Dialog Action} & \multicolumn{2}{c|}{Dialog Topic} & \multicolumn{2}{c}{Dialog Action} & \multicolumn{2}{c}{Dialog Topic} \\
                       & Acc.              & Bi.Acc.           & Acc.              & Bi.Acc.           & Acc.              & Bi.Acc.           & Acc.             & Bi.Acc.           \\ \midrule
MGCG                   & 84.78            & 86.52            & 64.31            & 66.65            & 85.23            & 88.05            & 57.68           & 58.29            \\
KERS                   & 89.17            & 90.49            & 76.34            & 79.33            & 86.15            & 88.75            & 63.18           & 65.06            \\
BERT                   & 90.19            & 91.35            & 83.53            & 85.61            & 91.32            & 93.26            & 66.87           & 67.77            \\
GPT-2                  & 91.76            & 93.03            & 86.24            & 87.47            & 91.64            & 93.88            & 65.63           & 66.80             \\
TPC                    & 93.58            & 95.11            & 91.92            & 93.53            & 93.21            & 94.62            & 83.26           & 84.49            \\ 
\midrule
FF-BPSN                & \textbf{97.11}            & \textbf{98.39}            & \textbf{92.97}            & \textbf{94.37}            & \textbf{97.68}            & \textbf{98.88}            & \textbf{93.15}           & \textbf{94.29}            \\ \bottomrule
\end{tabular}}
\caption{Planning results. \textbf{Bold} highlights the best results.}
\label{Table3}
\end{table}

\subsection{Datasets and Evaluation Metrics}
We evaluate our approach on DuRecDial \cite{liu-etal-2020-towards-conversational} (Chinese) and DuRecDial 2.0 \cite{liu2021durecdial} (English), both of which contain multi-turn dialogues incorporating domain knowledge and user profiles to support proactive target-oriented interactions. We adopt the version annotated by \cite{wang2023target}, and summarize the dataset details in Table~\ref{Table1}. Following prior work, response generation is assessed using \textbf{word-level F1} (F1), \textbf{BLEU-1/2}, \textbf{distinct} (DIST-1/2), \textbf{knowledge-F1} (Know. F1) and \textbf{target success} (Succ.). Dialogue path planning is evaluated using \textbf{accuracy} (Acc.) and \textbf{bigram accuracy} (Bi. Acc.).

\subsection{Baselines}
Our baselines for response generation include: \textbf{MGCG\_G} \cite{liu-etal-2020-towards-conversational}, which guides system dialogue generation based on the next action and topic; \textbf{KERS} \cite{zhang2021kers}, which enhances recommender dialogue systems with multiple subgoals; \textbf{TPC-BART} and \textbf{TPC-GPT} \cite{wang2023target}, which first plan a dialogue path and then fine-tune generation using BART and GPT-2, respectively; \textbf{BART} \cite{lewis2020bart}, an encoder-decoder pre-trained language model for text generation; \textbf{GPT-2} \cite{radford2019language}, an autoregressive pre-trained model for language generation; \textbf{DialoGPT} \cite{zhang2020dialogpt} (Chinese version \textbf{CDial-GPT} \cite{wang2020large}), pre-trained on large-scale dialogue corpora; and \textbf{LLaMA-3} \cite{DBLP:journals/corr/abs-2407-21783}, a widely used family of large language models. We experiment with two versions: LLaMA-1B and LLaMA-3B. For dialogue path planning, our baselines include: \textbf{MGCG} \cite{liu-etal-2020-towards-conversational}, which employs convolutional networks to predict the next action and topic; \textbf{KERS} \cite{zhang2021kers}, which uses transformer-based networks to predict the next action and topic; \textbf{BERT} \cite{devlin2019bert}, an encoder for predicting the next action and topic; \textbf{GPT-2} \cite{radford2019language}, a decoder for generating the next action and topic; and \textbf{TPC} \cite{wang2023target}, a target-oriented backward path planning model.

\subsection{Implementation Details}
During FF-BPSN training, we use AdamW with a learning rate of $1 \times 10^{-5}$, together with warmup and gradient clipping. FF-BPSN is trained for 20 epochs with a batch size of 4. The hidden layer dimension is set to 768, and both $\gamma$ and $\beta$ are set to 0.5. For fine-tuning the PLMs (DialoGPT, BART, GPT-2), we use a learning rate of $1 \times 10^{-5}$, train for 20 epochs, and set the batch size to 8. For the LLaMA series, models are fine-tuned using LoRA with rank 8 and alpha 32, a batch size of 2, and a learning rate of $1 \times 10^{-4}$ for 10 epochs. All experiments are conducted on GeForce RTX 4090 and RTX 3090 GPUs.

\subsection{Results and Analysis}

Table~\ref{Table2} reports response generation results on DuRecDial and DuRecDial 2.0. Here, FF denotes FF-BPSN, and w/ indicates “with.” LLaMA-3B with FF achieves the best performance in most metrics, including F1, BLEU, K.F1, and Succ., demonstrating its effectiveness in leveraging knowledge, generating natural transitions, and accomplishing targets. 

On DuRecDial, except for slightly lower DIST-1/2, all other metrics reach state-of-the-art performance. Models based on BART (TPC-BART and BART w/ FF) achieve the highest DIST-1/2 scores, indicating more diverse vocabulary usage. Incorporating FF leads to substantial gains across all language models; for example, Know. F1 reaches 57.34\%, reflecting improved knowledge utilization, and Succ. reaches 86.4\%, highlighting enhanced proactiveness and target achievement. LLaMA-1B performs slightly worse than LLaMA-3B but consistently outperforms GPT-2, suggesting overall performance scales with model size. Similar trends are observed on DuRecDial 2.0, with LLaMA-1B achieving competitive DIST-1/2 compared with TPC-BART, and LLaMA-3B w/ FF achieving 67.27\% on Know. F1 and 79.8\% on Succ., indicating stronger knowledge utilization and proactiveness. Overall, these results demonstrate that FF significantly enhances language model proactiveness and overall performance.

\begin{figure}[t]
    \centering
    \includegraphics[width=0.4\textwidth]{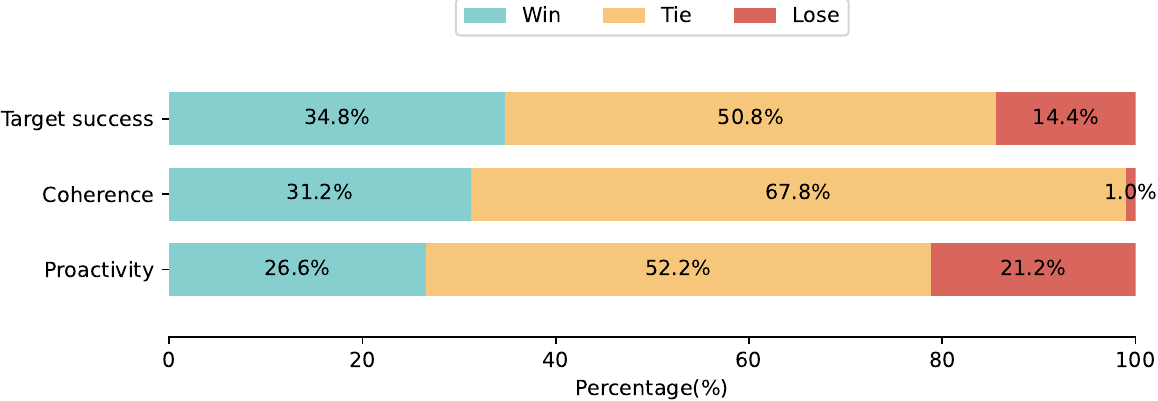}
    \caption{
    Human evaluation results.
    }
    \label{figure_human}
\end{figure}

\subsection{Human Evaluation}
We perform pairwise human evaluation on DuRecDial, comparing \textbf{LLaMA-3B} with and without FF-BPSN. A total of 500 responses are randomly selected, with results shown in Figure~\ref{figure_human}. The labels “win,” “tie,” and “lose” indicate cases where LLaMA-3B with FF-BPSN performs better, equally, or worse than LLaMA-3B, respectively. Evaluation is conducted along three dimensions: (i) \textit{proactivity}—whether responses demonstrate initiative and guidance; (ii) \textit{coherence}—whether responses are contextually appropriate; and (iii) \textit{target success}—whether responses achieve the target. We select three graduate students as evaluators, who are blinded to the correspondence between the models and their representations.

Compared to LLaMA-3B without FF-BPSN, human evaluation demonstrates that incorporating FF-BPSN yields substantial improvements, consistent with automatic metrics. In particular, LLaMA-3B with FF-BPSN significantly outperforms its counterpart in target success and proactiveness, demonstrating enhanced ability to guide users toward predefined targets. For coherence, LLaMA-3B with FF-BPSN achieves a high win rate, indicating better knowledge utilization and more context-consistent responses. To sum up, these results, aligned with automated evaluations, confirm that FF-BPSN substantially improves the performance of dialogue systems.

\subsection{Evaluation of Dialogue Path Planning}

To evaluate the effectiveness of the proposed path planning framework, we compare it with several baseline planning models, as shown in Table~\ref{Table3}. The results indicate that action recognition generally outperforms topic recognition, likely because the smaller set of actions is easier to identify, whereas the larger set of topics is more challenging. On DuRecDial, our framework achieves superior performance, particularly in action recognition. Compared with TPC, it improves topic recognition by 1.05\% in Acc. and 0.84\% in Bi. Acc. On DuRecDial 2.0, our framework shows notable gains across all metrics, boosting both action and topic recognition above 90\%. For example, topic Acc. and Bi. Acc. improve by 9.89\% and 9.8\%, respectively. These results demonstrate the effectiveness of FF-BPSN in dialogue path planning, highlighting its ability to generate reasonable paths that support improved response generation.

\begin{table}[!t]
\small
\renewcommand{\arraystretch}{0.7}
\centering
\setlength{\tabcolsep}{0.5mm}{
\begin{tabular}{lcccc}
\toprule
\multirow{2}{*}{Model} & \multicolumn{2}{c}{Dialog Action} & \multicolumn{2}{c}{Dialog Topic} \\
                       & Acc.       & Bi.Acc.       & Acc.       & Bi.Acc.      \\ \midrule
FF-BPSN$_{L=6}$                & 95.84           & 97.44               & 91.34           & 92.93              \\
FF-BPSN$_{L=8}$                & 95.50            & 96.95               & 89.61           & 91.18              \\
FF-BPSN$_{L=10}$               & 97.02           & \textbf{98.49}               & 92.40           & 93.97              \\
 \midrule
(Ours) FF-BPSN$_{L=12}$               & \textbf{97.11}           & 98.39               & \textbf{92.97}           & \textbf{94.37}              \\
OB BPSN$_{L=12}$            & 95.01           & 96.52               & 91.07           & 92.47              \\
OF BPSN$_{L=12}$            & 94.59           & 96.35               & 90.40            & 91.68              \\
BF BPSN$_{L=12}$            & 91.89           & 95.27               & 89.67           & 91.81              \\
FF-BPSN$_{L=12}$ w/o FF                 & 95.32           & 96.84               & 91.96           & 93.32              \\  \bottomrule
\end{tabular}}
\caption{Ablation study results.}
\label{Table4}
\end{table}

\subsection{Ablation Study}

To assess the effectiveness of the components introduced in this work, we conduct an ablation study with the following model variants: (1) \textbf{FF-BPSN$_{L=\{6,8,10,12\}}$}, representing different numbers of layers, with $L=12$ used in this study; (2) \textbf{OB}, using only the backward path; (3) \textbf{OF}, using only the forward path; (4) \textbf{BF}, backward-focused, where bidirectional path information is merged into the final backward path (forward-focused is used in this study); (5) \textbf{w/o FF}, without the forward-focused module. Results in Table~\ref{Table4} show that all components contribute to performance improvements, with deeper layers yielding better results, although even 6 layers remain competitive. Unidirectional planning (OF or OB) underperforms bidirectional planning, highlighting the importance of considering both directions. The forward-focused module provides the largest gains, whereas the backward-focused module is less effective. Overall, each component positively impacts path planning performance, with the best results achieved when bidirectional information is leveraged and forward information is emphasized.

\section{Conclusion}
We present FF-BPSN, a dialogue path planning framework for target-oriented proactive dialogue systems. FF-BPSN integrates two transformer-based encoders with a forward-focused module to perform bidirectional path planning, and the resulting path is used to guide language models in response generation. Extensive experiments demonstrate that FF-BPSN achieves state-of-the-art performance in path planning and substantially improves dialogue system effectiveness.

\vfill\pagebreak

\bibliographystyle{IEEEbib}
\bibliography{strings,refs}

@inproceedings{yang2022topkg,
  title={Topkg: Target-oriented dialog via global planning on knowledge graph},
  author={Yang, Zhitong and Wang, Bo and Zhou, Jinfeng and Tan, Yue and Zhao, Dongming and Huang, Kun and He, Ruifang and Hou, Yuexian},
  booktitle={Proceedings of the 29th International Conference on Computational Linguistics},
  pages={745--755},
  year={2022}
}

@inproceedings{ijcai2023p738,
  title={A Survey on Proactive Dialogue Systems: Problems, Methods, and Prospects},
  author={Deng, Yang and Lei, Wenqiang and Lam, Wai and Chua, Tat-Seng},
  booktitle={IJCAI},
  year={2023}
}

@article{wang2023target,
  title={A target-driven planning approach for goal-directed dialog systems},
  author={Wang, Jian and Lin, Dongding and Li, Wenjie},
  journal={IEEE Transactions on Neural Networks and Learning Systems},
  volume={35},
  number={8},
  pages={10475--10487},
  year={2023},
  publisher={IEEE}
}

@inproceedings{ma-etal-2021-cr,
    title = "{CR}-Walker: Tree-Structured Graph Reasoning and Dialog Acts for Conversational Recommendation",
    author = "Ma, Wenchang  and
      Takanobu, Ryuichi  and
      Huang, Minlie",
    editor = "Moens, Marie-Francine  and
      Huang, Xuanjing  and
      Specia, Lucia  and
      Yih, Scott Wen-tau",
    booktitle = "Proceedings of the 2021 Conference on Empirical Methods in Natural Language Processing",
    month = nov,
    year = "2021",
    address = "Online and Punta Cana, Dominican Republic",
    publisher = "Association for Computational Linguistics",
    url = "https://aclanthology.org/2021.emnlp-main.139/",
    doi = "10.18653/v1/2021.emnlp-main.139",
    pages = "1839--1851"
}

@inproceedings{liu-etal-2020-towards-conversational,
  title={Towards Conversational Recommendation over Multi-Type Dialogs},
  author={Liu, Z and Wang, H and Niu, ZYu and Wu, Hua and Che, W and Liu, T},
  booktitle={58th Annual Meeting of the Association-for-Computational-Linguistics (ACL) Conference Location ELECTR NETWORK},
  pages={1036--1049},
  year={2020},
  organization={ASSOC COMPUTATIONAL LINGUISTICS-ACL Location STROUDSBURG}
}

@inproceedings{wang-etal-2023-dialogue,
  title={Dialogue Planning via Brownian Bridge Stochastic Process for Goal-directed Proactive Dialogue},
  author={Wang, Jian and Lin, Dongding and Li, Wenjie},
  booktitle={The 61st Annual Meeting Of The Association For Computational Linguistics},
  year={2023}
}

@inproceedings{tang-etal-2019-target,
    title = "Target-Guided Open-Domain Conversation",
    author = "Tang, Jianheng  and
      Zhao, Tiancheng  and
      Xiong, Chenyan  and
      Liang, Xiaodan  and
      Xing, Eric  and
      Hu, Zhiting",
    editor = "Korhonen, Anna  and
      Traum, David  and
      M\`arquez, Llu\'\i s",
    booktitle = "Proceedings of the 57th Annual Meeting of the Association for Computational Linguistics",
    month = jul,
    year = "2019",
    address = "Florence, Italy",
    publisher = "Association for Computational Linguistics",
    url = "https://aclanthology.org/P19-1565/",
    doi = "10.18653/v1/P19-1565",
    pages = "5624--5634"
}

@inproceedings{zhong2021keyword,
  title={Keyword-guided neural conversational model},
  author={Zhong, Peixiang and Liu, Yong and Wang, Hao and Miao, Chunyan},
  booktitle={Proceedings of the AAAI Conference on Artificial Intelligence},
  volume={35},
  number={16},
  pages={14568--14576},
  year={2021}
}

@inproceedings{liu2020gochat,
  title={Gochat: Goal-oriented chatbots with hierarchical reinforcement learning},
  author={Liu, Jianfeng and Pan, Feiyang and Luo, Ling},
  booktitle={Proceedings of the 43rd International ACM SIGIR Conference on Research and Development in Information Retrieval},
  pages={1793--1796},
  year={2020}
}

@inproceedings{xu2020knowledge,
  title={Knowledge graph grounded goal planning for open-domain conversation generation},
  author={Xu, Jun and Wang, Haifeng and Niu, Zhengyu and Wu, Hua and Che, Wanxiang},
  booktitle={Proceedings of the AAAI conference on artificial intelligence},
  volume={34},
  number={05},
  pages={9338--9345},
  year={2020}
}

@inproceedings{zhang2021kers,
  title={KERS: A knowledge-enhanced framework for recommendation dialog systems with multiple subgoals},
  author={Zhang, Jun and Yang, Yan and Chen, Chencai and He, Liang and Yu, Zhou},
  booktitle={Findings of the Association for Computational Linguistics: EMNLP 2021},
  pages={1092--1101},
  year={2021}
}

@inproceedings{liu2021durecdial,
  title={DuRecDial 2.0: A Bilingual Parallel Corpus for Conversational Recommendation},
  author={Liu, Zeming and Wang, Haifeng and Niu, Zheng-Yu and Wu, Hua and Che, Wanxiang},
  booktitle={Proceedings of the 2021 Conference on Empirical Methods in Natural Language Processing},
  pages={4335--4347},
  year={2021}
}

@article{radford2019language,
  title={Language models are unsupervised multitask learners},
  author={Radford, Alec and Wu, Jeffrey and Child, Rewon and Luan, David and Amodei, Dario and Sutskever, Ilya and others},
  journal={OpenAI blog},
  volume={1},
  number={8},
  pages={9},
  year={2019}
}

@article{wang2024target,
  title={Target-constrained Bidirectional Planning for Generation of Target-oriented Proactive Dialogue},
  author={Wang, Jian and Lin, Dongding and Li, Wenjie},
  journal={ACM Transactions on Information Systems},
  volume={42},
  number={5},
  pages={1--27},
  year={2024},
  publisher={ACM New York, NY}
}

@article{deng2025proactive,
  title={Proactive Conversational AI: A Comprehensive Survey of Advancements and Opportunities},
  author={Deng, Yang and Liao, Lizi and Lei, Wenqiang and Yang, Grace Hui and Lam, Wai and Chua, Tat-Seng},
  journal={ACM Transactions on Information Systems},
  volume={43},
  number={3},
  pages={1--45},
  year={2025},
  publisher={ACM New York, NY}
}

@inproceedings{dong-etal-2025-protod,
  title={Protod: Proactive task-oriented dialogue system based on large language model},
  author={Dong, Wenjie and Chen, Sirong and Yang, Yan},
  booktitle={Proceedings of the 31st International Conference on Computational Linguistics},
  pages={9147--9164},
  year={2025}
}

@inproceedings{lin2024screen,
  title={SCREEN: A Benchmark for Situated Conversational Recommendation},
  author={Lin, Dongding and Wang, Jian and Leong, Chak Tou and Li, Wenjie},
  booktitle={Proceedings of the 32nd ACM International Conference on Multimedia},
  pages={9591--9600},
  year={2024}
}

@inproceedings{li-etal-2025-multi-hop,
    title = "Multi-Hop Question Generation via Dual-Perspective Keyword Guidance",
    author = "Li, Maodong  and
      Zhang, Longyin  and
      Kong, Fang",
    editor = "Che, Wanxiang  and
      Nabende, Joyce  and
      Shutova, Ekaterina  and
      Pilehvar, Mohammad Taher",
    booktitle = "Findings of the Association for Computational Linguistics: ACL 2025",
    month = jul,
    year = "2025",
    address = "Vienna, Austria",
    publisher = "Association for Computational Linguistics",
    url = "https://aclanthology.org/2025.findings-acl.526/",
    doi = "10.18653/v1/2025.findings-acl.526",
    pages = "10096--10112",
    ISBN = "979-8-89176-256-5"
}

@article{DBLP:journals/corr/abs-2407-21783,
	publtype={informal},
	author={Abhimanyu Dubey and Abhinav Jauhri and Abhinav Pandey and et al.},
	title={The Llama 3 Herd of Models},
	year={2024},
	cdate={1704067200000},
	journal={CoRR},
	volume={abs/2407.21783},
	url={https://doi.org/10.48550/arXiv.2407.21783}
}

@inproceedings{devlin2019bert,
  title={Bert: Pre-training of deep bidirectional transformers for language understanding},
  author={Devlin, Jacob and Chang, Ming-Wei and Lee, Kenton and Toutanova, Kristina},
  booktitle={Proceedings of the 2019 conference of the North American chapter of the association for computational linguistics: human language technologies, volume 1 (long and short papers)},
  pages={4171--4186},
  year={2019}
}

@inproceedings{lewis2020bart,
  title={BART: Denoising Sequence-to-Sequence Pre-training for Natural Language Generation, Translation, and Comprehension},
  author={Lewis, Mike and Liu, Yinhan and Goyal, Naman and Ghazvininejad, Marjan and Mohamed, Abdelrahman and Levy, Omer and Stoyanov, Veselin and Zettlemoyer, Luke},
  booktitle={Proceedings of the 58th Annual Meeting of the Association for Computational Linguistics},
  pages={7871--7880},
  year={2020}
}

@inproceedings{wang2020large,
  title={A large-scale chinese short-text conversation dataset},
  author={Wang, Yida and Ke, Pei and Zheng, Yinhe and Huang, Kaili and Jiang, Yong and Zhu, Xiaoyan and Huang, Minlie},
  booktitle={CCF International Conference on Natural Language Processing and Chinese Computing},
  pages={91--103},
  year={2020},
  organization={Springer}
}

@inproceedings{zhang2020dialogpt,
  title={DIALOGPT: Large-Scale Generative Pre-training for Conversational Response Generation},
  author={Zhang, Yizhe and Sun, Siqi and Galley, Michel and Chen, Yen-Chun and Brockett, Chris and Gao, Xiang and Gao, Jianfeng and Liu, Jingjing and Dolan, William B},
  booktitle={Proceedings of the 58th ACL},
  pages={270--278},
  year={2020}
}

\end{document}